\documentclass[runningheads]{llncs}

\usepackage{eccv}

\usepackage{eccvabbrv}

\usepackage{graphicx}
\usepackage{booktabs}
\usepackage{multirow}
\usepackage{float}
\usepackage{bbding}
\usepackage{hyperref} 
\usepackage[accsupp]{axessibility}  

\usepackage{hyperref}

\usepackage{orcidlink}

\begin{document}

\title{Hybrid-LUT: Channel-Aware Hybrid Lookup Table and Filtering for Efficient Image Denoising} 

\titlerunning{Abbreviated paper title}

\author{Zhilin Ai\inst{1}*\orcidlink{0009-0009-6082-809X} \and Boyu Li\inst{1}*\orcidlink{0000-0001-9709-9673} \and
Sidi Yang\inst{1}\orcidlink{0009-0002-9769-854X} \and
Wenqing Shi\inst{1}\orcidlink{0009-0005-6890-7113} \and
Wenyong Zhou\inst{1}\orcidlink{0009-0008-0427-7935} \and
Binxiao Huang\inst{1}\orcidlink{0000-0001-5316-703X} \and
Chenchen Ding\inst{1} \and
Ngai Wong\inst{1}\Envelope}

\authorrunning{F.~Author et al.}

\institute{The University of Hong Kong\\
\email{\{zhilin.ai,u3011228,mennyshi,wenyongz,huangbx7,dingcc\}@connect.hku.hk, \{liboyu,nwong\}@eee.hku.hk}}

\begingroup
\renewcommand{\thefootnote}{}
\footnotetext{* Contributed equally \\
\Envelope\ Corresponding author}
\endgroup

\maketitle

\begin{abstract}
Lookup table (LUT)-based image denoising methods have attracted increasing attention due to their high efficiency and hardware-friendly properties. However, existing RGB-LUT approaches require three identical LUTs to process RGB channels in parallel, resulting in large on-chip SRAM consumption. A simple alternative is to apply LUT processing only to the luminance (Y) channel in the YUV color space to reduce memory usage. However, this naive strategy leads to degraded restoration quality, since ignoring the chrominance (UV) channels introduces color distortion and residual artifacts. In this work, we propose Hybrid-LUT, a YUV-based asymmetric channel-processing framework that combines LUT and filtering in a unified design. Specifically, a multi-band LUT branch with pixel-level weight fusion is applied to the Y channel to recover fine textures, while lightweight filtering is used for the UV channels to maintain color consistency. This design reduces LUT storage by two-thirds compared with RGB-LUT methods while maintaining the same runtime throughput. Extensive experiments show that Hybrid-LUT achieves state-of-the-art (SOTA) performance across multiple benchmarks with only 421 KB of storage. In particular, our method surpasses existing LUT-based denoising approaches by at least 0.63 dB CPSNR on real-world datasets, demonstrating its effectiveness for image denoising on resource-constrained edge devices. The project is available at \url{https://github.com/Ai-ZL/Hybrid-LUT}.
\end{abstract}

\section{Introduction}
\label{sec:intro}

Image restoration aims to reconstruct high-quality (HQ) images from low-quality (LQ) inputs degraded by noise, compression artifacts, or color distortions. 
In recent years, deep neural networks (DNNs) have achieved remarkable progress in various image restoration tasks, including super-resolution, denoising, and deblurring. 
Despite their strong performance, these DNN-based methods typically require massive computational resources and high memory bandwidth, which limits their deployment on edge devices with constrained power and storage budgets. 
Consequently, increasing research efforts have shifted toward lightweight and hardware-friendly alternatives that maintain competitive performance while significantly reducing computational complexity.

Lookup table (LUT)-based approaches have recently emerged as a promising solution due to their ``space-for-time'' trade-off mechanism. 
Instead of performing intensive floating-point operations, LUT methods pre-compute the mappings between degraded inputs and restored outputs and store them in compact tables for direct indexing during inference. 
This strategy completely eliminates multiply–accumulate operations and enables deterministic hardware behavior with extremely low latency~\cite{huang2024HKLUT,li2025bdlutsid, Ma2022SPLUT, liboyu2024}. 
Despite these advantages, LUT-based methods suffer from a fundamental limitation: the storage cost grows exponentially with input dimensionality. 
For a 3D RGB LUT with $N$ quantization levels, the required storage scales as $\mathcal{O}(N^3)$, which quickly exceeds the on-chip SRAM capacity available in modern hardware systems. 
Although several compression or cascaded LUT designs have been proposed to alleviate this issue, such as multi-stage or diagonal-first compression strategies~\cite{li2024SPFLUT}, the redundancy caused by processing three RGB channels independently remains largely unresolved. 
Even when all channels share the same LUT structure, practical hardware implementations still require three parallel lookup pipelines to maintain throughput, leading to substantial memory consumption.

To address this limitation, we propose a novel \textbf{Hybrid-LUT} framework that integrates LUT-based and filter-based processing within the YUV color space. 
Benefiting from the decorrelation property of YUV, the proposed framework decomposes the input RGB image into luminance (Y) and chrominance (U, V) components. 
The Y channel, which carries most of the structural and intensity information, is enhanced using a dedicated LUT mapping, while the UV channels are refined using lightweight filtering operations due to their lower dynamic range, reduced spatial sensitivity, and relatively smooth characteristics. 
This asymmetric processing strategy effectively reduces LUT storage by approximately two-thirds compared with conventional RGB-LUT approaches, while maintaining identical runtime throughput.
Moreover, by incorporating only a lightweight mean filtering operation with negligible hardware overhead, the proposed method achieves denoising performance that surpasses SOTA approaches.

Furthermore, we design a novel complementary multi-band LUT architecture with pixel-level weight fusion to overcome the limitations of static and content-agnostic LUT mappings (e.g., fixed kernel stacking or cascaded structures). 
Specifically, we exploit bit-plane characteristics to construct separate LUT branches for the Most Significant Bits (MSBs) and Least Significant Bits (LSBs), enabling the model to capture diverse frequency components. 
A variance-guided fusion mechanism is then employed to dynamically balance edge-preserving and smoothing operations at the pixel level, allowing the framework to suppress noise adaptively while preserving fine structural details.

Extensive experiments on multiple benchmark datasets demonstrate the superiority of the proposed \textbf{Hybrid-LUT} framework. 
Our method achieves state-of-the-art (SOTA) restoration performance while requiring only \textbf{421~KB} of LUT storage, highlighting its strong practicality for real hardware deployment. 
It achieves comparable or superior PSNR to existing LUT-based approaches while maintaining only one-third of the memory footprint and preserving real-time inference speed. 
The main contributions of this work are summarized as follows:


\begin{itemize}

\item We propose the first asymmetric channel-processing scheme that strategically combines LUT and filtering in YUV space for denoising. By performing LUT-based restoration only on the structure-rich Y channel in the YUV space while applying lightweight filtering to the UV channels (U, V), our design effectively eliminates redundant RGB lookups. This hybrid design reduces on-chip SRAM storage by $66.7\%$ compared with conventional RGB-LUT schemes while maintaining identical runtime throughput, making it highly suitable for practical hardware deployment.

\item We introduce a novel multi-band LUT branch with pixel-level weight fusion to overcome the limitations of static and content-agnostic LUT mappings. By decomposing the Y signal into MSB and LSB branches and adaptively blending their outputs across bit-planes, the proposed method captures diverse frequency components, enabling effective noise suppression while preserving fine textures.

\item Hybrid-LUT achieves SOTA restoration performance with extreme hardware efficiency. On the real-world SIDD dataset, our method surpasses existing LUT-based approaches by at least 0.63 dB CPSNR while requiring only 421 KB of storage. Compared with DNN-based methods, it provides over $196\times$ speedup and consumes only $0.1\%$ of the energy, demonstrating strong potential for real-time FPGA/ASIC implementations.

\end{itemize}

\section{Related Works}
Image restoration methods fall into three paradigms: deep learning–based, traditional filter–based, and LUT–based. As Fig.~\ref{fig:hybridlut_intro} shows, they differ in restoration quality, inference speed, model size, and memory footprint. DNNs excel in fidelity but are costly; filters are simple yet often mediocre; LUTs enable fast inference at the expense of storage. This motivates our hybrid design, which balances these trade‑offs, as detailed later.

\subsection{Deep Learning--Based Image Restoration}

Deep neural networks (DNNs) have achieved remarkable success in image restoration tasks, including denoising, deblurring, and image enhancement. Early works such as DnCNN~\cite{Zhang2017DnCNN} and U-Net variants employ convolutional architectures to learn end-to-end mappings from degraded inputs to restored outputs. More recent approaches, including SwinIR~\cite{Liang2021SwinIR} and Restormer~\cite{waqas2022restormer}, further improve restoration performance by introducing hierarchical attention mechanisms and transformer-based architectures. 

Despite their strong representation capability, these models typically rely on large numbers of parameters and extensive floating-point operations, resulting in high latency and significant power consumption in practical hardware deployment. Although lightweight variants and FPGA-oriented accelerators, such as Light-DnCNN and CNN-based implementations, have been proposed to reduce model complexity, they still demand considerable logic resources and memory bandwidth. Consequently, achieving real-time image restoration under strict resource and power constraints remains challenging for purely learning-based approaches.

{
\setlength{\textfloatsep}{4pt}   
\setlength{\floatsep}{4pt}        
\begin{figure}[!t]
\setlength{\belowcaptionskip}{-0.3cm}
\centering
\includegraphics[scale = 0.11]{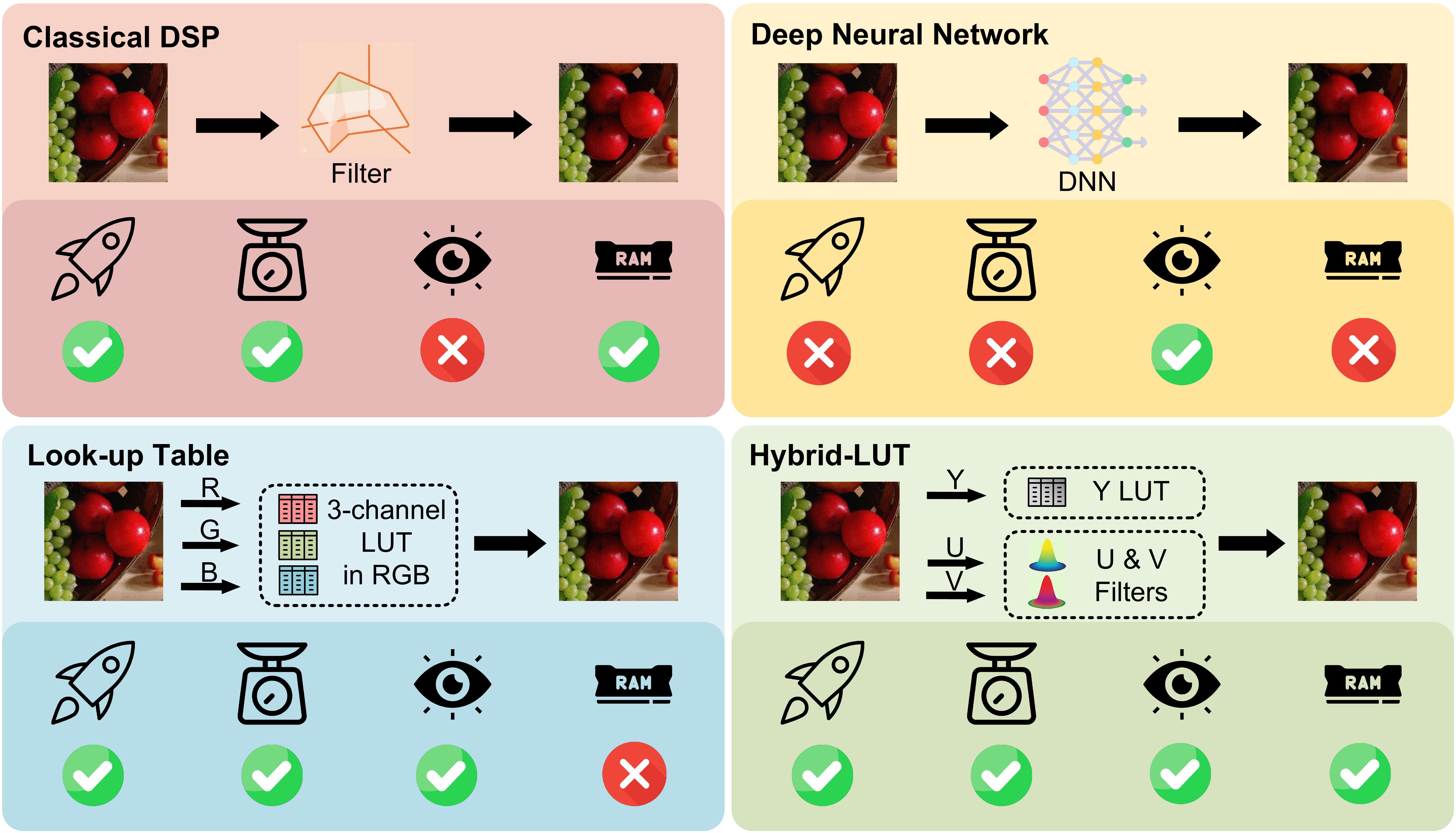}
\caption{Denoising schemes. Top-left: Classical method (e.g., CBM3D); top-right: Deep Neural Network (e.g., DnCNN); bottom-left: Look-up Table (i.e., conventional RGB-LUT); bottom-right: our Hybrid-LUT. The icons denote speed (rocket), model size (scale), visual effects (eyes), and physical memory consumption (RAM).}
\label{fig:hybridlut_intro}
\end{figure}
}

\subsection{Traditional Filter--Based Methods}

Before the advent of deep learning, image restoration was primarily addressed using hand-crafted filtering or optimization-based techniques. Classical filters such as the bilateral filter (BF) and its joint variants smooth images while preserving edges by adaptively weighting pixel similarities in both spatial and intensity domains~\cite{Tomasi1998Bi,Durand2002Bilateral}. Other approaches, including weighted least-squares (WLS) filtering and guided filtering~\cite{he2010guided}, further enhance edge preservation and visual quality.

However, these methods are often computationally demanding due to iterative optimization procedures or high-order convolution operations. Although several approximations and hardware-accelerated implementations of BF have been proposed~\cite{Adams2010FastBF, Spagnolo2023De}, they still require substantial arithmetic operations and memory accesses. As a result, while filter-based methods offer favorable interpretability and decent restoration quality, their scalability and efficiency remain limited, particularly for deployment on low-power edge devices.

\subsection{LUT--Based Methods}

To alleviate the heavy computational cost of both DNN-based and filtering-based approaches, lookup table (LUT) methods have recently emerged as an efficient alternative~\cite{Jo2021SRLUT, Li2022MuLUT, Liu2023RCLUT}. By precomputing the mappings between degraded inputs and restored outputs, LUT methods enable near-zero-computation inference through simple table indexing. 

Early channel-wise 3D LUT techniques were widely used for efficient color enhancement, while spatial LUT frameworks such as SRLUT~\cite{Jo2021SRLUT} and MuLUT~\cite{Li2022MuLUT} extended this concept to patch-based image super-resolution. Subsequent works, including SPLUT~\cite{Ma2022SPLUT}, RCLUT~\cite{Liu2023RCLUT}, HKLUT~\cite{huang2024HKLUT}, and BDLUT~\cite{li2025bdlutsid}, further enlarged receptive fields or aggregated cascaded LUT modules to improve restoration accuracy while preserving real-time inference speed. Additionally, to mitigate the memory increase from stacking LUTs to enlarge the receptive field, SPFLUT~\cite{li2024SPFLUT} introduces a diagonal-first compression strategy, while DNLUT~\cite{yang2025dnlut} balances storage and performance through joint cross-channel/spatial processing and a specialized L‑shaped kernel.

Nevertheless, most existing LUT-based methods process RGB channels independently or identically, which introduces substantial redundancy in storage and computation. More importantly, such designs ignore the perceptual differences between Y and UV components, which play distinct roles in human visual perception and color image restoration.

Existing studies have largely optimized learning-based, filtering-based, and LUT-based pipelines in isolation. However, the redundant RGB lookups in LUT methods and the high computational cost of full-channel filtering remain unresolved. Motivated by these limitations, we explore a hybrid strategy in the YUV color space that decomposes RGB images into Y and UV components. Our approach leverages LUT-based processing for the structure-rich Y channel while employing lightweight filtering for the UV channels. This hybrid design effectively reduces hardware storage requirements while maintaining high restoration fidelity, bridging the gap between LUT efficiency and filter adaptivity.

{
\setlength{\textfloatsep}{4pt}   
\setlength{\floatsep}{4pt}        
\begin{figure}[!t]
\setlength{\belowcaptionskip}{-0.3cm}
  \centering
  \includegraphics[width=1\textwidth]{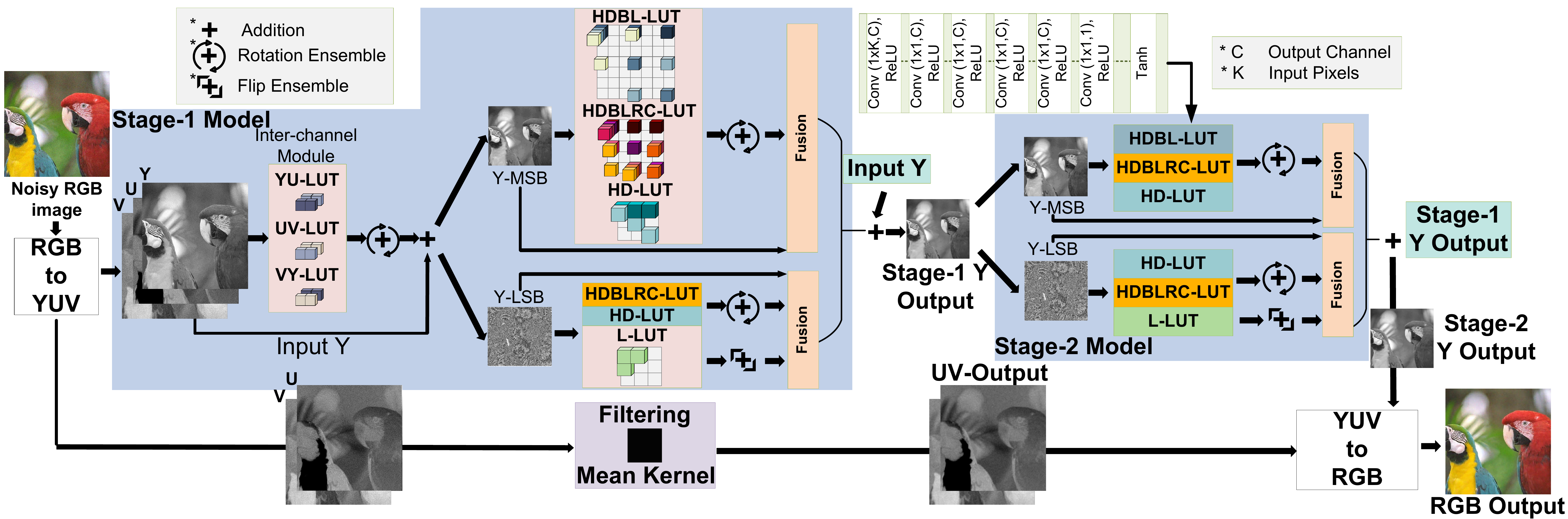}
  \caption{Overall architecture of Hybrid-LUT: After YUV cross-channel processing, the Y channel is split into MSB and LSB branches, each with three parallel LUT units. The second stage output is combined with the UV mean-filter results.}
  \label{fig:LUT_framework}
\end{figure}
}

  
  
  

\section{Method}

\subsection{Preliminary}
SRLUT~\cite{Jo2021SRLUT} pioneered storing CNN traversal outputs in a LUT, using a finite receptive field (RF), sampled inputs, and interpolation to recover CNN learning capabilities with drastically reduced resources. Later, MuLUT~\cite{Li2022MuLUT} expanded the RF via multiple kernel designs, while SPLUT~\cite{Ma2022SPLUT} and HKLUT~\cite{huang2024HKLUT} split INT8 inputs into 4‑bit MSB and LSB branches, slightly sacrificing accuracy to eliminate interpolation and reduce latency. The recent DNLUT~\cite{yang2025dnlut} adopts an INT8+interpolation structure with a Pairwise Channel Mixer (PCM) to achieve SOTA denoising.

Although LUTs trade storage for computation, existing RGB‑LUT methods must replicate tables across channels for parallel hardware processing, causing memory to triple. Moreover, in RGB space, structural information and noise are uniformly distributed, forcing identical operations on each channel and exacerbating redundancy. In contrast, YUV space inherently decouples Y from UV. Because the Y channel concentrates most structural information and is critical for perceptual clarity, we assign a powerful LUT to it; the U and V channels, having lower dynamic range and less spatial detail, are processed with lightweight filters. This asymmetric, perceptually‑aligned design achieves quality comparable to full RGB processing at a fraction of the hardware memory cost.

\begin{table}[H]
\renewcommand{\arraystretch}{0.85}
\setlength{\belowcaptionskip}{-0.005cm}
\caption{Effect of separate Y and UV processing on CBSD68 denoising (CPSNR/dB, AWGN $\sigma=15$). GT: ground truth; Y: SRLUT; UV: Mean Filter.}
\centering{
\setlength{\tabcolsep}{1.8mm}
\small
\begin{tabular}{ccc}
\hline
Y Method & UV Method & CPSNR (dB) \\ \hline
SRLUT & Mean Filter & 31.40\\
 SRLUT & GT & 32.15  \textcolor{blue}{(+0.75)}\\
 GT & Mean Filter & 39.12 \textcolor{blue}{(+7.72)}\\
  \hline
\end{tabular}}
\label{tab:y_uv_compare}
\end{table}
Table \ref{tab:y_uv_compare} provides an upper‑bound analysis by replacing denoised channels with ground truth (GT). Upgrading UV from a mean filter to GT yields only 0.75 dB gain when Y is fixed, whereas upgrading Y from SRLUT to GT gives a 7.72 dB improvement when UV is fixed. This confirms that the Y channel dominates restoration quality, while investing complexity in UV brings less returns—motivating our asymmetric design.

\subsection{Overview of the Proposed Framework}
To address the aforementioned issues, we propose the Hybrid-LUT with three key innovative structures: 1) asymmetric channel processing architecture for Y and UV channels; 2) hierarchical feature extraction based on weighted fusion; and 3) kernel structure designs with complementary frequency bands.

\noindent\textbf{Training Network.} In Fig.~\ref{fig:LUT_framework}, we construct an asymmetric YUV denoising framework. The Y channel, carrying primary visual texture, is processed by a high-precision hierarchical LUT fusion mechanism, while the smoother UV channels undergo lightweight mean filtering. After nonlinear gain amplification for energy normalization, cross-channel features are extracted via an inter-channel module derived from~\cite{yang2025dnlut}. The 8-bit Y channel is decomposed into 4-bit MSB and 4-bit LSB components, each processed by three complementary LUT units. Feature reconstruction employs flip/rotation ensembles and spatially adaptive weighted fusion, with a two-stage cascaded residual architecture ensuring high performance.

The LUT module for each kernel sampling pattern corresponds to a six-layer CNN during training. For MSB/LSB branches, the first layer uses $1\times3$ kernels, followed by four $1\times1$ conv layers with 64 channels and ReLU, and a final $1\times1$ layer with $ReLU+Tanh$. The inter-channel module adopts dense connectivity: a $1\times4$ $conv+GELU$, four dense blocks ($1\times1$ $conv+ReLU$ with concatenation), and a $1\times1$ $conv+Tanh$.

 \noindent\textbf{Transferring to LUT. }For the 3/4 input elements per kernel, the CNN is transformed into a 3/4D-LUT. We enumerate all possible input combinations to compute the corresponding CNN outputs, which are stored in the LUT using the inputs as indices. The LUT storage requirement is $v^n$, where $v$ represents the number of quantization levels and n is the input dimension. The input range $[0, 255]$ is uniformly sampled using $2^4$ sampling interval, and the outputs are rounded to 8-bit integers in the range $[-127, 127]$.

 \noindent\textbf{LUT Test. }During inference, networks are replaced by LUT lookups. A branch's output $\hat{y_i}$ is computed as:
 \begin{small} 
\begin{equation}\label{eq_lut}
\hat{y_i}=\frac{1}{N}\sum_{k=0}^{N}\left(\frac{1}{M_k}\sum_{j=0}^{M_k}R_j^{-1}(LUT_k(R_j(x_i)))\right)
\end{equation}
\end{small} 
where $x_i$ denotes input pixels,$LUT_k$ represents the $k^{th}$ lookup table, $R_j$ and $R_j^{-1}$ implement the $j^{th}$ 90° rotation and its inverse operation (with $M_k=4/2$ rotations per kernel), and $N$ is the kernel count.

\subsection{Weighted LUTs Fusion}
Current LUT-based denoising methods\cite{huang2024HKLUT,Li2022MuLUT,Ma2022SPLUT,li2024SPFLUT} primarily enhance performance by expanding the RF through multi-stage cascades or stacking kernels with diverse sampling patterns. However, these approaches rely on static, content-agnostic logic that fails to adapt to local structures. Furthermore, as the number of kernels increases, the linear combination of mapping units acts as a low-pass filter, resulting in excessive smoothing of edges and the loss of fine texture details.

To address this limitation, we introduce an adaptive fusion mechanism based on statistical priors for the MSB and LSB branches. First, a $5 \times 5$ sliding window computes the local variance $\sigma^2$ as a measure of texture complexity. This normalized variance serves as an index to rapidly look up fusion weights from a lightweight 1-D LUT, thereby enabling adaptive behavior. These weights modulate the outputs of multiple LUT units, each with distinct spectral characteristics: in texture-rich regions (high variance), the model automatically amplifies edge-preserving units to maintain sharpness; in flat regions (low variance), it prioritizes units with strong noise suppression capabilities to eliminate color artifacts and residual noise. Through this design, our method achieves an optimal balance between denoising strength and detail preservation, overcoming the limitations of static structural feature representations.

{
\setlength{\textfloatsep}{4pt}   
\setlength{\floatsep}{4pt}        
\begin{figure}[!t]
\setlength{\belowcaptionskip}{-0.5cm}
\centering
\includegraphics[width=8.5cm]{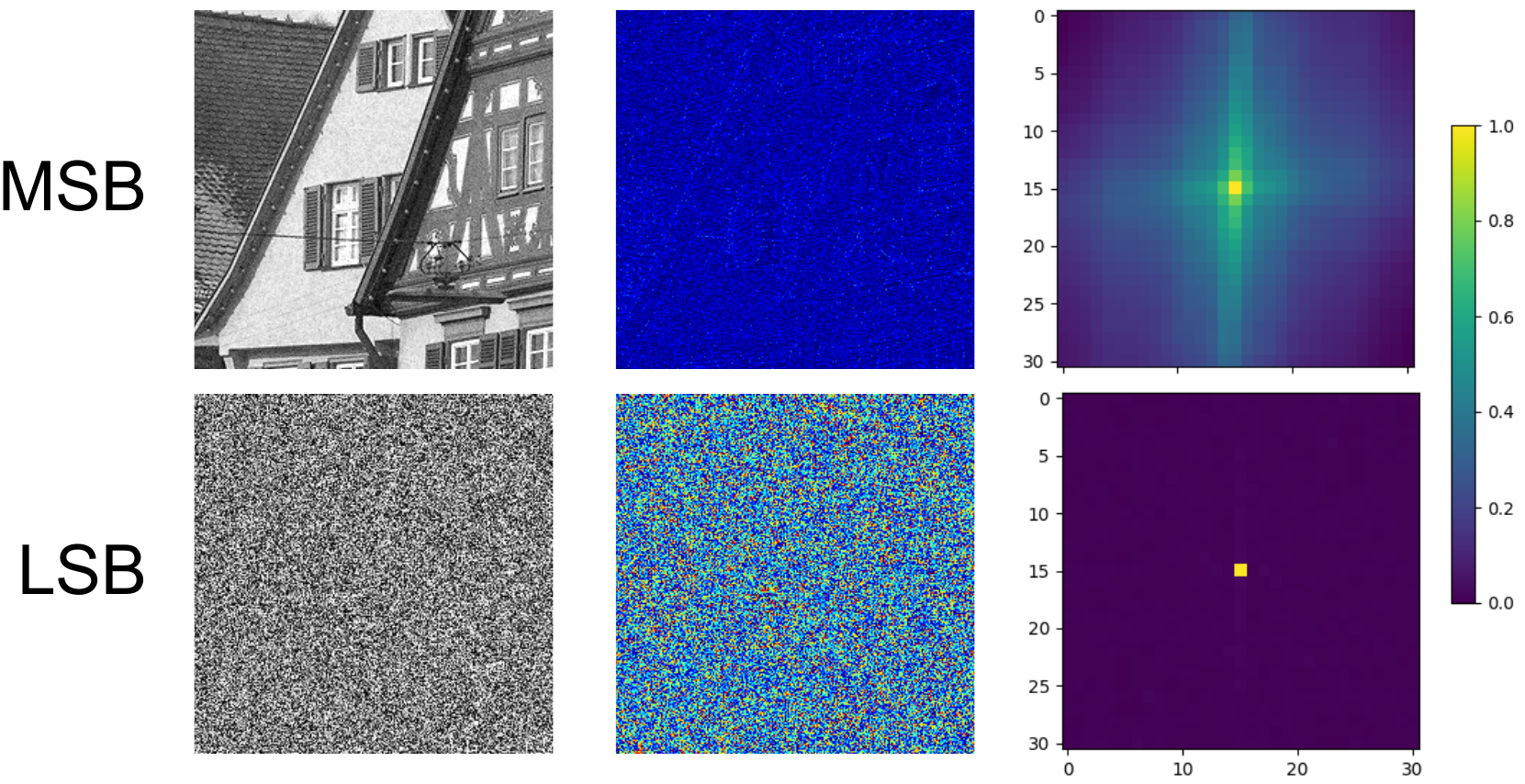}
\caption{Statistical Differences between MSB and LSB Bit-Planes. Left: visualization of noisy MSB and LSB; middle: noise residual heatmaps; right: spatial autocorrelation plots.}
\label{fig: msb_lsb}
\end{figure}
}

\subsection{Complementary Kernel Designs}

The intrinsic limitation of a single LUT lies in its restricted input dimensionality (typically 3–4 pixels), which inherently limits its RF. To overcome this without exponentially increasing the table size, we propose a multi-band LUT ensemble within the MSB/LSB branch based on complementary kernel patterns. The MSB and LSB branches each employ multiple LUT units targeting distinct frequency bands, with different unit combinations per branch.

We design four 3-pixel kernel combination structures using rotation ensembles and dilation to achieve varying RFs (HD-LUT, HDBLRC-LUT, HDBL-LUT, L-LUT as in Fig. \ref{fig:LUT_framework}). The ensemble rotates the input by a set of angles: four rotations ($0^{\circ}$, $90^{\circ}$, $180^{\circ}$, $270^{\circ}$) for the first three kernel combinations, and two ($0^{\circ}$, $180^{\circ}$) for the last. Specifically, HD-LUT covers a $5\times5$ region using horizontal/diagonal kernels; HDBLRC‑LUT ($d=1$) covers $9\times9$ with six dilated (d=1) kernels (including horizontal, diagonal, bird, top-left L, bottom-right L, bottom-left L); HDBL‑LUT ($d=2$) covers $13\times13$ with four dilated (d=2) kernels (including horizontal, diagonal, bird, top-left L); and L‑LUT covers $3\times3$ with a single L‑shaped kernel.

For the MSB branch, which carries the primary structural information, its visualization (Fig. \ref{fig: msb_lsb} Left) reveals clear object contours and topological skeletons. Crucially, its spatial autocorrelation plot (Fig. \ref{fig: msb_lsb} Right) exhibits a broad, slowly decaying peak, indicating that signals (and their parasitic noises) in the MSB are spatially correlated over a wider range. To handle such structural dependencies, we employ a dilation-heavy ensemble: HD-LUT, HDBLRC-LUT ($d=1$), and HDBL-LUT ($d=2$). While the star-shaped HD unit focuses on the immediate 8-connectivity neighborhood to preserve sharp edges, the HDBLRC and HDBL units significantly expand the RF through dilated sampling to maintain color consistency in flat areas. This combination allows the MSB branch to suppress heavy noise while preserving a robust structural skeleton.

In contrast, the LSB branch is dominated by stochastic noise, with its signal nearly submerged in quantization-level residuals (Fig. \ref{fig: msb_lsb} Middle). Its autocorrelation plot (Fig. \ref{fig: msb_lsb} Right) exhibits a sharp, impulse-like spike, confirming highly localized information with no long-range dependency. Accordingly, we adopt a localized ensemble: HD-LUT, HDBLRC-LUT ($d=1$), and L-LUT. The L-LUT prioritizes the current pixel and its immediate neighbors to capture high-frequency residuals with maximum fidelity. Using a $d=2$ dilated unit (HDBL) on the LSB would introduce irrelevant, uncorrelated noise from distant pixels, leading to the blurring of fine textures. The shift from HDBL in MSB to L-Unit in LSB reflects a transition from ``structural denoising'' to ``fine-detail refinement''.

By selecting different kernel combinations, our design aligns the RF with the physical characteristics of each bit-plane and implements a coarse-to-fine processing logic. Long-range dependencies are handled in the MSB branch, while computational resources focus on pixel-level precision in the LSB branch. The resulting heterogeneous ensemble enables effective noise suppression across the entire frequency spectrum. Moreover, this diversity mitigates the over-smoothing effect common in kernel stacking, as the dynamic fusion mechanism can adaptively weight the most suitable LUT based on the local texture complexity.

{
\setlength{\textfloatsep}{4pt}   
\setlength{\floatsep}{4pt}        
\begin{figure}[!t]
  \centering
  \setlength{\belowcaptionskip}{-0.3cm}
  \begin{subfigure}{0.48\textwidth}
    \centering
    \includegraphics[width=\textwidth]{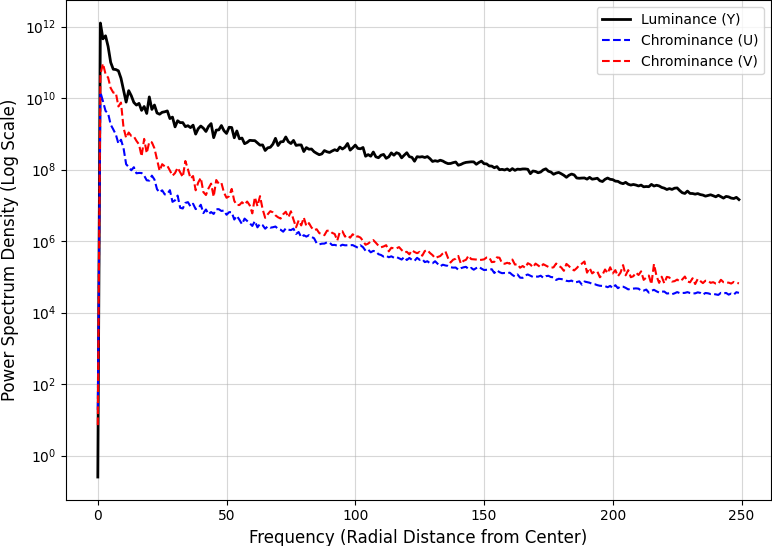}
    \caption{Power Spectral Density: Y vs. UV}
    \label{fig:clean_yuv_spc}
  \end{subfigure}
  \hfill
  \begin{subfigure}{0.48\textwidth}
    \centering
    \includegraphics[width=\textwidth]{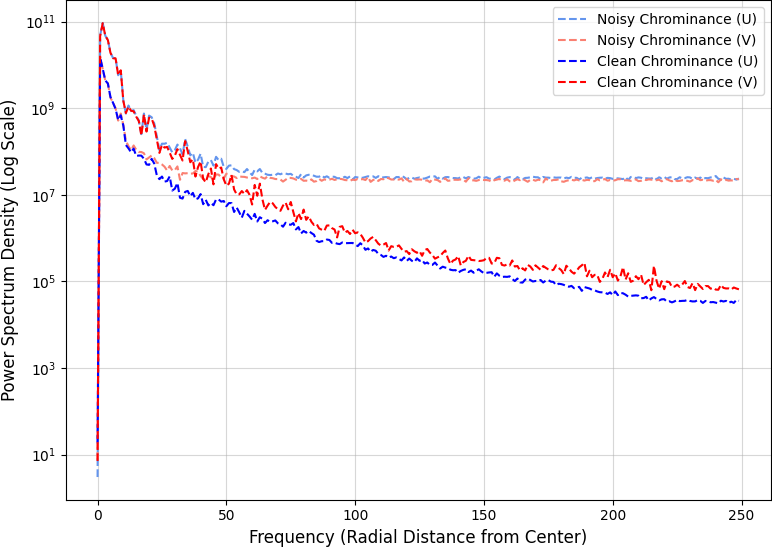}
    \caption{Power Spectral Density: Clean vs. Noisy UV}
    \label{fig:noisy_uv_clean_uv_spc}
  \end{subfigure}
  \caption{Power Spectral Density (PSD) of YUV channels. (a): PSD curves of the Y, U, and V channels for the clean image. (b): PSD curves of the U and V channels for both the clean image and its noisy version.}
  \label{fig:YUV_SPC}
\end{figure}
}

\subsection{Asymmetric Channel Processing Architecture}

To achieve an optimal trade-off between restoration quality and computational overhead, we propose an Asymmetric Channel Processing Architecture. This design is grounded in the distinct frequency-domain characteristics of Y and UV channels.

We used Power Spectral Density (PSD) analysis~\cite{Burton:87}. As shown in Fig. \ref{fig:clean_yuv_spc}, the clean Y channel retains significantly higher energy in the high-frequency regime compared to the UV channels, indicating that structural details are predominantly stored in the Y component. In contrast, the spectrum of clean UV channels decays rapidly, confirming their inherently low-bandwidth nature. Crucially, as shown in Fig. \ref{fig:noisy_uv_clean_uv_spc}, the energy gap between clean and noisy UV channels is concentrated in the high-frequency bands. This implies that high-frequency components in the noisy UV channels are dominated by stochastic noise rather than meaningful textures.

Based on these observations, we employ a high-precision LUT architecture for the Y channel, where both the MSB and LSB branches are each composed of multiple complementary LUT units to recover delicate textures. For the UV channels, we substitute the complex LUT with a lightweight Mean Filter. Since the UV signal is naturally smooth, a spatial low-pass filter can effectively suppress high-frequency color artifacts with negligible loss of visual fidelity. This asymmetric strategy allocates most of the storage and computing resources to the Y channel through LUT operations, reducing memory footprint by two-thirds compared to conventional parallel RGB-LUT architecture, effectively maximizing perceptual gain.

\setlength{\textfloatsep}{4pt}   
\setlength{\floatsep}{4pt}        

\begin{table}[t]
\setlength{\belowcaptionskip}{-0.005cm}
\renewcommand{\arraystretch}{0.7}  
\caption{Size and Quantitative comparison (CPSNR/dB) for color image denoising on 4 fixed-noise-intensity benchmark datasets. Ours(Y) denotes asymmetric YUV processing (LUT on Y, mean filter on UV); Ours(YUV) denotes symmetric processing with LUTs on all three channels.}
\centering
\tiny  
\setlength{\tabcolsep}{2pt}  
\renewcommand{\arraystretch}{1}
\begin{tabular}{@{}c|c|c|c|ccc|ccc|ccc|ccc@{}}
\hline
\multirow{2}{*}{\textbf{Cat.}} & \multirow{2}{*}{\textbf{Method}} & \multicolumn{2}{c|}{\textbf{Size}} & \multicolumn{3}{c|}{\textbf{CBSD68}($\sigma$)} & \multicolumn{3}{c|}{\textbf{Kodak24}($\sigma$)} & \multicolumn{3}{c|}{\textbf{Urban100}($\sigma$)} & \multicolumn{3}{c@{}}{\textbf{McMaster}($\sigma$)} \\ 
\cline{3-4} \cline{5-7} \cline{8-10} \cline{11-13} \cline{14-16}
 & & \textbf{(KB)} & \textbf{(MB)} & \textbf{15} & \textbf{25} & \textbf{50} & \textbf{15} & \textbf{25} & \textbf{50} & \textbf{15} & \textbf{25} & \textbf{50} & \textbf{15} & \textbf{25} & \textbf{50} \\ \hline
 \multirow{7}{*}{LUT} & SRLUT\cite{Jo2021SRLUT} & 82 & 1.312 & 29.76 & 26.71 & 22.41 & 30.35 & 27.16 & 22.65 & 29.38 & 26.04 & 21.60 & 31.18 & 28.01 & 23.35 \\
  & BDLUT\cite{li2025bdlut} & 66 & 1.056 & 30.36 & 27.78 & 24.45 & 31.07 & 28.52 & 24.99 & 29.96 & 27.05 & 23.30 & 31.98 & 29.44 & 25.80 \\
 & MuLUT\cite{Li2022MuLUT} & 490 & 7.84 & 30.52 & 28.11 & 24.85 & 31.31 & 29.02 & 25.28 & 30.25 & 27.67 & 23.75 & 32.28 & 29.88 & 26.36 \\
  & RCLUT\cite{Liu2023RCLUT} & 326 & 5.216 & 30.68 & 28.12 & 25.04 & 31.57 & 29.07 & 25.89 & 30.33 & 27.80 & 23.86 & 32.51 & 29.89 & 26.50 \\
 & SPFLUT\cite{li2024SPFLUT} & 618 & 9.888 & 30.97 & 28.56 & 25.33 & 31.86 & 29.58 & 26.26 & 30.89 & 28.26 & 24.22 & 31.77 & 30.44 & 26.91 \\
  & DNLUT\cite{yang2025dnlut} & 518 & 5.384 & 32.41 & 29.88 & 26.03 & 33.02 & 30.24 & 26.74 & 32.12 & 28.87 & 25.01 & 32.88 & 30.44 & 27.12 \\
 & \textbf{Ours(Y)} & 421 & 1.684 & 32.47 & 29.72 & 26.32 & 32.78 & 30.15 & 26.85 & 31.17 & 28.46 & 25.00 & 31.78 & 29.40 & 26.06 \\
  & \textbf{Ours(YUV)} & 421 & 3.796 & 32.67 & 29.80 & 26.48 & 33.11 & 30.39 & 27.19 & 31.75 & 28.87 & 25.35 & 32.92 & 30.17 & 26.96 \\ \hline
 \multirow{2}{*}{Class.} & CBM3D\cite{Dabov2007CBM3D} & — & — & 33.52 & 30.71 & 27.38 & 34.28 & 31.68 & 29.02 & 33.92 & 31.35 & 27.94 & 34.06 & 31.66 & 28.51 \\
 & MCWNNM\cite{Xu2017MCWNNM} & — & — & 31.98 & 29.32 & 26.98 & 33.23 & 30.89 & 28.67 & 30.23 & 29.23 & 27.00 & 31.23 & 30.20 & 27.55 \\ \hline
 \multirow{2}{*}{DNN} & DnCNN\cite{Zhang2017DnCNN} & 2.65MB & — & 33.90 & 31.23 & 27.95 & 34.60 & 32.14 &  28.96 & 32.98 & 30.81 & 27.59 & 33.45 & 31.52 & 28.62 \\
 & SwinIR\cite{Liang2021SwinIR} & 117MB & — & 34.42 & 31.78 & 28.56 & 35.34 & 32.89 & 29.79 & 35.61 & 33.20 & 30.22 & 35.13 & 32.90 & 29.82 \\ \hline
\end{tabular}
\label{tab:color_denosing}
\end{table}

\begin{table}[t]
\setlength{\belowcaptionskip}{-0.005cm}
\renewcommand{\arraystretch}{0.75}
\caption{Quantitative comparison (CPSNR/SSIM) on real-world color image denoising. For DnD, only PSNR is available via online evaluation.}
\centering
\setlength{\tabcolsep}{1pt}
\renewcommand{\arraystretch}{1}
\tiny  
\resizebox{\columnwidth}{!}{
\begin{tabular}{c|c|ccccccccc}
\hline
Data. & Method & SRLUT\cite{Jo2021SRLUT} & MuLUT\cite{Li2022MuLUT}& BDLUT\cite{li2025bdlut} & SPFLUT\cite{li2024SPFLUT}  & DNLUT\cite{yang2025dnlut} & Ours(Y) & Ours(YUV) & CBM3D\cite{Dabov2007CBM3D} & DnCNN\cite{Zhang2017DnCNN}\\ \hline
 \multirow{2}{*}{SIDD} & CPSNR & 29.38 & 33.24 & 33.18 & 34.91 & 35.44 & 36.07 & 35.62 & 30.14 & 36.45\\
 & SSIM & 0.634 & 0.830 & 0.924 & 0.865 & 0.875 &  0.934 & 0.911 & 0.702 & 0.900\\
  \hline
 \multirow{2}{*}{DnD} & PSNR & 33.39 & 35.11 & 35.25 & 36.22 & 36.67 & 36.82 & 35.58 & 33.12 & 37.11\\
 & SSIM  & 0.839 & 0.868 & 0.892 & 0.911 & 0.922 & 0.928 & 0.916 & 0.823 & 0.932\\
  \hline
\end{tabular}
}
\label{tab:color_denosing_real}
\end{table}

\section{Experiments}
\label{sec:experiments}
\subsection{Experiment Setup}
\textbf{Datasets and Metrics. }We use DIV2K\cite{Agustsson_2017_DIV2K} as the training set. For evaluation, we employ four benchmark datasets: CBSD68\cite{Martin2001cbsd}, Kodak24\cite{Rich1999kodak}, McMaster\cite{Lei2022McMaster}, and Urban100\cite{Huang2015urban}. Both training and testing datasets are corrupted with Additive White Gaussian Noise (AWGN) at noise levels of $\sigma = 15, 25, 50$ to assess the framework's performance on known noise patterns. To evaluate real-world noise handling capability, we use the SIDD training dataset\cite{abdelrahman2018SIDD} for training, with SIDD validation and DnD datasets\cite{tobias2017dnd} serving as benchmarks. We adopt color peak signal-to-noise
ratio(CPSNR) and structural similarity index(SSIM) as evaluation metrics to assess denoising effectiveness across color channels and structural/perceptual quality, respectively.

\textbf{Experimental Setting. }The network is trained for 200K iterations with a batch size of 16 on Nvidia RTX 3090 GPUs. The Adam optimizer($\beta_1= 0.9, \beta_2=0.999$ and $\epsilon=1e-8$) with the MSE loss is used on the Y channel. We employ a cosine annealing schedule that starts the learning rate at $1\times10^{-4}$ and decays it smoothly to $5\times10^{-5}$ over the entire training duration. The YUV image is chosen as input and is randomly cropped into $48 \times 48$ patches, and the dataset is enhanced by random rotation and flipping. To minimize indexing artifacts, final outputs utilize 4D simplex interpolation, 1D linear interpolation, and softmax. During the validation phase, Y-PSNR is adopted as the evaluation metric to monitor the restoration performance of the LUT model. The parameter of the UV Mean Filter is grid-searched to maximize CPSNR.

\textbf{Baselines. } We evaluate LUT framework against several SOTA denoising methods and LUT-based methods, including CBM3D\cite{Dabov2007CBM3D}, MCWNNM\cite{Xu2017MCWNNM}, DnCNN\cite{Zhang2017DnCNN}, SwinIR\cite{Liang2021SwinIR}, SRLUT\cite{Jo2021SRLUT}, BDLUT\cite{li2025bdlut}, MuLUT\cite{Li2022MuLUT}, RCLUT\cite{Liu2023RCLUT}, SPFLUT\cite{li2024SPFLUT}, DNLUT\cite{yang2025dnlut}.

{
\setlength{\textfloatsep}{4pt}   
\setlength{\floatsep}{4pt}        
\begin{figure}[!t]
\setlength{\belowcaptionskip}{-0.3cm}
\centering
\includegraphics[scale = 0.19]{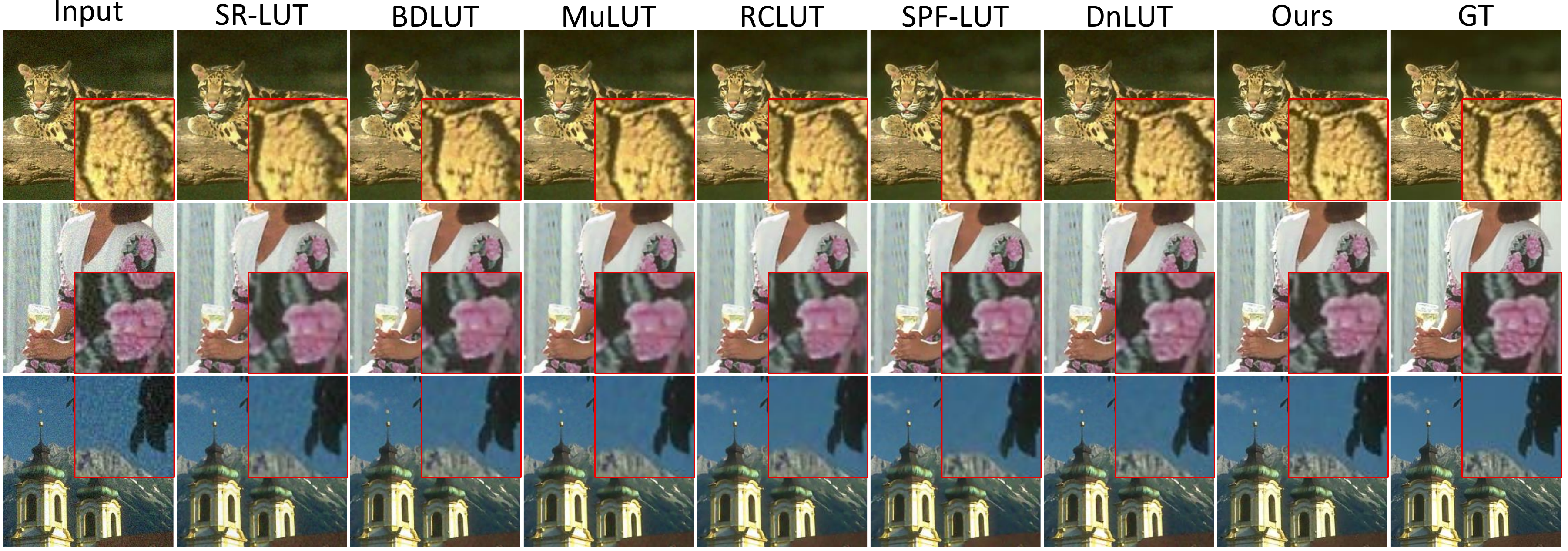}
\caption{Qualitative evaluation of color image denoising ($\sigma = 15$ AWGN) performance on CBSD68 dataset. }
\label{fig:gaussian_comparison}
\end{figure}

\begin{figure}[!t]
\setlength{\belowcaptionskip}{-0.3cm}
\centering
\includegraphics[scale = 0.19]{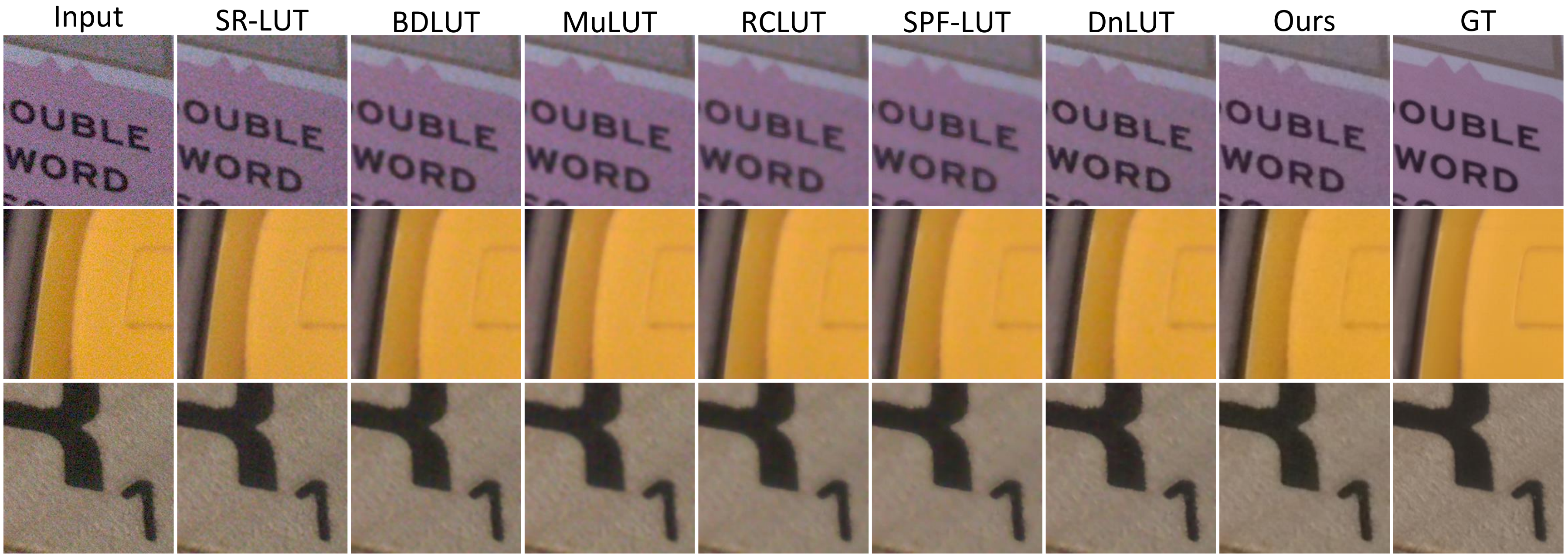}
\caption{Qualitative evaluation of color image denoising (Real-world noise) performance on SIDD dataset. }
\label{fig:sidd_comparison}
\end{figure}
}

\subsection{Quantitative results}
Table \ref{tab:color_denosing} compares our asymmetric architecture Ours(Y) with other common RGB architecture LUT-based methods and other well-known denoising methods. Unlike RGB architectures that require separate LUT storage for each channel, our approach stores LUTs only for the Y channel, reducing hardware memory requirements to one-third while achieving significant savings. The two size columns represent: (left) model size and (right) physical memory consumption when deploying multi-channel, multi-rotation kernels on hardware platforms. Our approach demonstrates substantial memory efficiency advantages while maintaining competitive CPSNR performance. While SRLUT and BDLUT occupy the smallest memory footprints, their performance is constrained by their original design goals---super-resolution and blind denoising, respectively. In contrast, our method, specifically optimized for denoising, creates a significant performance gap. Our method achieves an average CPSNR improvement of over 1.35 dB across benchmarks. Compared to MuLUT with a slightly larger model size, our method achieves an average improvement of 0.91 dB across benchmarks. Against the current SOTA DNLUT, our method (Ours(Y)) is highly competitive: it outperforms DNLUT on CBSD68 and Kodak24 while delivering comparable results on complex scenes such as Urban100—all with nearly $30\%$ of the physical memory required by DNLUT. These results demonstrate that by concentrating complex LUT-based reconstruction on the Y channel and applying lightweight filtering to UV, our architecture achieves competitive denoising accuracy at an extremely low resource cost. Furthermore, our symmetric architecture Ours(YUV) achieves performance comparable to or better than DNLUT across most datasets, with slightly lower storage consumption.

Table \ref{tab:color_denosing_real} presents a quantitative evaluation of our proposed asymmetric YUV architecture on the SIDD and DnD datasets, demonstrating its exceptional capacity for real-world color image denoising. Within the category of LUT-based methods, Ours(Y) consistently outperforms the current SOTA DNLUT by 0.63 dB (35.44 vs 36.07 dB) in CPSNR, and also surpasses our symmetric architecture Ours(YUV) by 0.45 dB, while maintaining a substantially lower memory footprint. Notably, our ultra-lightweight solution achieves performance competitive with the deep learning-based DnCNN; on the SIDD dataset, Ours(Y) even surpasses DnCNN in terms of SSIM (0.934 vs. 0.900), demonstrating superior structural integrity and texture fidelity. Furthermore, our method achieves a substantial 5.93 dB gain over the classical CBM3D algorithm on SIDD, demonstrating its robustness against complex, sensor-dependent noise distributions. These results collectively validate that our Y-focused asymmetric design achieves an optimal balance between hardware efficiency and restoration quality, rivaling far more computationally expensive DNNs in practical scenarios.

\subsection{Qualitative results}

Fig. \ref{fig:gaussian_comparison} and \ref{fig:sidd_comparison} visually demonstrate the superior denoising capability of our proposed asymmetric YUV architecture on both synthetic AWGN ($\sigma=15$) and real-world noisy images. As shown in Fig. \ref{fig:gaussian_comparison}, existing LUT-based methods tend to oversmooth textured regions while failing to remove blotchy artifacts in flat areas. In contrast, our method achieves cleaner backgrounds with better color fidelity (e.g., the sky in the third row) and preserves fine textures more effectively, such as the pattern of a cheetah's coat in the first row and the texture of the dress in the second row, yielding results visually closest to the GT. Similarly, Fig. \ref{fig:sidd_comparison} further validates the advantages of our approach on real-world noise. Compared to all other LUT-based methods, our results exhibit clearer text and wood grain textures, more uniform color patches consistent with the GT.

{
\setlength{\textfloatsep}{0pt}   
\setlength{\floatsep}{0pt}        
\begin{table}[t]
\setlength{\belowcaptionskip}{-0.005cm}
\renewcommand{\arraystretch}{1.08}
\caption{Denoising comparison (CPSNR/dB) of channel processing architectures on four color benchmark datasets ($\sigma=15$ AWGN).}
\centering{
\setlength{\tabcolsep}{2pt}
\footnotesize  
\begin{tabular}{c|c|c|c|c|cccc}
\hline
 Y & UV & R & GB & Physical Memory & CBSD68& Kodak24& Urban100& McMaster\\ \hline
  LUT & Filter & - & - & 1684 KB & 32.47 & 32.78 & 31.17 & 31.78 \\
  LUT & LUT & - & - & 3796 KB & 32.67 & 33.11 & 31.75 & 32.92 \\
   - & - & LUT & Filter & 1684 KB & 28.11 & 28.83 & 25.79 & 30.17 \\
  \hline
\end{tabular}}
\label{tab:yuv_rgb_ablation}
\end{table}
}

{
\setlength{\textfloatsep}{4pt}   
\setlength{\floatsep}{4pt}        
\begin{table}[t]
\setlength{\belowcaptionskip}{-0.005cm}
\renewcommand{\arraystretch}{1.05}
\caption{Denoising comparison (CPSNR/dB) of weighted fusion configurations on four color benchmark datasets ($\sigma=15$ AWGN).}
\centering{
\setlength{\tabcolsep}{2pt}
\footnotesize  
\begin{tabular}{c|c|cccc}
\hline
 weight calculation & weight granularity & CBSD68& Kodak24& Urban100& McMaster\\ \hline
fixed average & per-tensor & 32.42 & 32.71 & 31.11 & 31.58 \\
  learnable & per-tensor & 32.36 & 32.65 & 31.06 & 31.57 \\
   learnable & per-row & 32.39 & 32.67 & 31.07 & 31.56 \\  
  learnable & pixel-wise & 32.47 & 32.78 & 31.17 & 31.78 \\
  \hline
\end{tabular}}
\label{tab:weight_ablation}
\end{table}

\begin{table}[t]
\setlength{\belowcaptionskip}{-0.005cm}
\renewcommand{\arraystretch}{0.95}
\caption{Denoising comparison (CPSNR/dB) of different LUT configurations for MSB and LSB branches on four color benchmark datasets ($\sigma=15$ AWGN).}
\centering{
\setlength{\tabcolsep}{2pt}
\tiny  
\begin{tabular}{c|c|c|cccc}
\hline
 MSB Method & LSB Method & Storage (KB) & CBSD68& Kodak24& Urban100& McMaster\\ \hline
  HD & HD & 278 & 32.17 & 32.43 & 31.38 & 30.83 \\
  HD/HDBL & HD/L & 321 & 32.41 & 32.70 & 31.05 & 31.72 \\
  HD/HDBLRC & HD/HDBLRC & 379 & 32.41 & 32.71 & 31.09 & 31.57 \\
   HDBL/HD/HDBLRC & L/HD/HDBLRC & 421 & 32.47 & 32.78 & 31.17 & 31.78 \\
   HDBL/HD/HDBLRC & HDBL/HD/HDBLRC & 446 & 32.47 & 32.76 & 31.11 & 31.75 \\
  \hline
\end{tabular}}
\label{tab:branch_ablation}
\end{table}
}

\subsection{Ablation study}
We conduct ablation studies on the color image denoising task with AWGN ($\sigma=15$) to evaluate the contribution of each module in Hybrid-LUT.


\noindent\textbf{YUV channel asymmetric architecture. }Table \ref{tab:yuv_rgb_ablation} ablates channel processing architectures.Comparing asymmetric YUV and RGB configurations, the YUV-based model outperforms its RGB counterpart by 1.61–5.38 dB in CPSNR. This substantial gap stems from the high inter-channel correlation in RGB space, where processing a single channel (R) cannot adequately restore shared structural information. In contrast, YUV decomposition concentrates structural content into the Y channel, enabling efficient high-fidelity restoration. In addition, the symmetric configuration improves CPSNR by 0.2-1.14 dB but triples memory consumption. This confirms that the Y channel carries the vast majority of perceptually significant information. The larger gap on McMaster likely reflects its richer color textures, which demand more precise UV reconstruction. Overall, our asymmetric approach achieves near-peak performance with minimal hardware overhead.

\noindent\textbf{Fusion weights. }Table \ref{tab:weight_ablation} evaluates fusion strategies for multiple LUT units across two dimensions: weight calculation method and granularity. First, as it progresses from coarse (a single weight shared across the entire tensor, i.e., per-tensor) to fine (pixel-wise), CPSNR consistently improves, with the pixel-wise weights fusion achieving the best performance across all benchmarks. Second, fixed-average fusion outperforms coarse-grained weights, highlighting the necessity of spatial adaptation. This indicates that global weights fail to account for the spatial heterogeneity of image textures and noise. Our proposed pixel-wise learnable fusion overcomes this limitation by enabling spatially-adaptive blending, dynamically modulating the contribution of each LUT branch based on local context to achieve superior denoising performance.

\noindent\textbf{LUT units performance. }Table \ref{tab:branch_ablation} investigates the impact of various LUT unit combinations on the MSB and LSB branches. Our final configuration (MSB: HDBL/HD/HDBLRC; LSB: L/HD/HDBLRC) consistently achieves the highest CPSNR across all four benchmarks. Notably, this design at 421 KB outperforms the more resource-intensive symmetric configuration at 446 KB (e.g., by 0.06 dB on Urban100), further confirming that adding large-RF units to the LSB branch introduces irrelevant noise. These results validate the complementarity of our specialized kernels and the effectiveness of our bit-plane-aware resource allocation, which prioritizes structural reconstruction in the MSB branch while maintaining efficiency in the LSB branch.

\subsection{Efficiency Evaluation}
Table \ref{tab:efficiency_eval} evaluates Hybrid-LUT on energy cost, runtime, and storage. Following AdderSR \cite{Song2020AdderSRTE}, we estimate energy by counting MAC operations. While our asymmetric design consumes more energy than DNLUT, it remains far more efficient than SPFLUT and DNN methods. Runtime measured on Android (standard Java API) shows that Hybrid‑LUT processes $512\times512$ images in a time similar to DNLUT's, occupying $5.3\%$ of SPFLUT's time and $0.35\%$ of DnCNN's; further gains are possible via FPGA. Storage‑wise, Hybrid‑LUT's model size is $421$ KB ($97$ KB smaller than DNLUT), and its single‑channel LUT reduces physical memory to $31.28\%$ of DNLUT's, enabling deployment on resource‑constrained devices.

{
\setlength{\textfloatsep}{4pt}   
\setlength{\floatsep}{4pt}        
\begin{table}[t]
\setlength{\belowcaptionskip}{-0.005cm}
\renewcommand{\arraystretch}{0.95}
\caption{Efficiency evaluation of color image denoising methods on mobile platform (Samsung Exynos 1580). Energy costs are measured on $512\times512$ color images.}
\centering{
\setlength{\tabcolsep}{2pt}
\tiny
\begin{tabular}{c|c|c|ccccc}
\hline
\multirow{2}{*}{Cat.} & \multirow{2}{*}{Method} & \multirow{2}{*}{Platform} & Runtime (ms) & Runtime (ms) & \multirow{2}{*}{Energy Cost (pJ)} & \multirow{2}{*}{Storage (KB)}& Actual\\
 &  &  & $256\times256$ & $512\times512$ &  & & Storage (MB)\\ \hline
 \multirow{7}{*}{LUT} & SRLUT\cite{Jo2021SRLUT} & Mobile & 24 & 73 & 149.98M & 82 & 1.312 \\
  & BDLUT\cite{li2025bdlut} & Mobile & 44 & 142 & 458.77M & 66 &  1.056 \\
 & MuLUT\cite{Li2022MuLUT} & Mobile & 76 & 281 & 899.88M & 490 & 7.84 \\
 & RCLUT\cite{Liu2023RCLUT} & Mobile & 66 & 259 & 612.92M & 326 & 5.216 \\
 & SPFLUT\cite{li2024SPFLUT} & Mobile & 1,962 & 7,576 & 2.32G & 30,178 & 482.848 \\
  & DNLUT\cite{yang2025dnlut} & Mobile & 103 & 403 & 687.34M & 518 & 5.384 \\
 & Ours(Y) & Mobile & 104 & 399 & 700.06M & 421 & 1.684 \\
  \hline
 \multirow{2}{*}{Classical} & CBM3D\cite{Dabov2007CBM3D} & PC & 8,197 & 35,808 & 4.82G & - & -  \\
 & MCWNNM\cite{Xu2017MCWNNM}  & PC & 151,256 & 2,640,250 & 89.23G & - & - \\
  \hline
\multirow{2}{*}{DNN} & DnCNN\cite{Zhang2017DnCNN} & Mobile & 20,397 & 115,633 & 542.53G & 2,239 & - \\
& SwinIR\cite{Liang2021SwinIR}  & Mobile & 511,724 & 3,122,750 & 12.03T & 45,499 & - \\
\hline
\end{tabular}}
\label{tab:efficiency_eval}
\end{table}
}

\section{Conclusion}
In this paper, we present Hybrid-LUT, a novel and efficient framework that pioneers the asymmetric channel-processing paradigm for LUT-based image denoising. By decoupling the restoration process into the YUV color space, our method effectively addresses the memory redundancy inherent in traditional RGB-LUT schemes. Specifically, we concentrate sophisticated multi-band LUT units and pixel-level weight fusion on the Y channel to recover intricate textures, while employing lightweight filtering for the UV channels. This strategic resource re-allocation allows Hybrid-LUT to reduce LUT storage by two-thirds compared to conventional architectures. Ultimately, with a small memory footprint, our method achieves highly competitive results with SOTA DNLUT and even outperforms it by 0.63 dB on the SIDD real-world dataset, demonstrating a favorable performance–storage trade-off. Given its superior restoration quality and extreme hardware efficiency, Hybrid-LUT serves as an ideal solution for high-fidelity image denoising on resource-constrained edge devices.


\section*{Acknowledgements}
This project is supported in part by the Theme-based Research Scheme (TRS) project T45-701/22-R and GRF Project 17203224 of the Research Grants Council (RGC), Hong Kong SAR, and in part by the AVNET-HKU Emerging Microelectronics \& Ubiquitous Systems (EMUS) Lab.

%
%
\bibliographystyle{splncs04}
\bibliography{main}

\end{document}